\documentclass[10pt,twocolumn,letterpaper]{article}

\usepackage[T1]{fontenc}
\usepackage[utf8]{inputenc}
\usepackage{authblk}

\usepackage{framed}
\usepackage{cvpr}
\usepackage{times}
\usepackage{epsfig}
\usepackage{graphicx}
\usepackage{amsmath}
\usepackage{amssymb}
\usepackage{multirow}
\usepackage{amsthm}
\usepackage{amsfonts}
\usepackage{bbm}

\usepackage[pagebackref=true,breaklinks=true,letterpaper=true,colorlinks,bookmarks=false]{hyperref}

\cvprfinalcopy 

\def\cvprPaperID{1445} 

\ifcvprfinal\fi
\begin{document}

\title{RNN Fisher Vectors for Action Recognition and Image Annotation}

\author[1,2]{Guy Lev}
\author[1]{Gil Sadeh}
\author[1]{Benjamin Klein}
\author[1]{Lior Wolf}
\affil[1]{The Blavatnik School of Computer Science, Tel Aviv University, Israel}
\affil[2]{IBM Research, Haifa, Israel}

\renewcommand\Authands{ and }

\maketitle

\begin{abstract}
Recurrent Neural Networks (RNNs) have had considerable success in classifying and predicting sequences. We demonstrate that RNNs can be effectively used in order to encode sequences and provide effective representations. The methodology we use is based on Fisher Vectors, where the RNNs are the generative probabilistic models and the partial derivatives are computed using backpropagation. State of the art results are obtained in two central but distant tasks, which both rely on sequences: video action recognition and image annotation. 
We also show a surprising transfer learning result from the task of image annotation to the task of video action recognition.
\end{abstract}

\section{Introduction}

Fisher Vectors have been shown to provide a significant performance gain on many different applications in the domain of computer vision ~\cite{simonyan2013fisher,peng2014action,chatfield2011devil,perronnin2010large}. In the domain of video action recognition, Fisher Vectors and Stacked Fisher Vectors~\cite{peng2014action} have recently outperformed state-of-the-art methods on multiple datasets~\cite{peng2014action,wang2015action}. Fisher Vectors (FV) have also recently been applied to word embedding (e.g. word2vec~\cite{mikolov2013distributed}) and have been shown to provide state of the art results on a variety of NLP tasks~\cite{lev2015defense}, as well as on image annotation and image search tasks~\cite{klein2015associating}.

In all of these contributions, the FV of a set of local descriptors is obtained as a sum of gradients of the log-likelihood of the descriptors in the set with respect to the parameters of a probabilistic mixture model that was fitted on a training set in an unsupervised manner. In spite of being richer than the mean vector pooling method, Fisher Vectors based on a probabilistic mixture model are invariant to order. This makes them less appealing for annotating, for example, video, in which the sequence of events determines much of the meaning.

This work presents a novel approach for FV representation of sequences using a Recurrent Neural Network (RNN). The RNN is trained to predict the next element of a sequence given the previous elements. Conveniently, the gradients needed for the computation of the FV are extracted using the available backpropagation infrastructure. 

The new representation is 
sensitive to ordering and therefore mitigates the disadvantage of using the standard Fisher Vector representation. It is applied to two different and challenging tasks: video action recognition and image annotation by sentences.

Several recent works have proposed to use an RNN for sentence representation~\cite{sutskever2014sequence,bahdanau2014neural,palangi2015deep,kiros2015skip}. The Recurrent Neural Network Fisher Vector (RNN-FV) method differs from these works in that a sequence is represented by using derived gradient from the RNN as features, instead of using a hidden or an output layer of the RNN. 

The paper explores two different approaches for training the RNN for the image annotation and image search tasks. In the classification approach, the RNN is trained to predict the following word in the sentence. The regression approach tries to predict the embedding of the following word (i.e. treating it as a regression task). The large vocabulary size makes the regression approach more scalable and achieves better results than the classification approach. In the video action recognition task, the regression approach is the only variant being used, since the notion of a discrete word does not exist. The VGG~\cite{oxford} Convolutional Neural Network (CNN) is used to extract features from the frames of the video and the RNN is trained to predict the embedding of the next frame given the previous ones. Similarly, C3D~\cite{tran2014learning} features of sequential video sub-volumes are used with the same training technique.   

Although the image annotation and video action recognition tasks are quite different, a surprising boost in performance in the video action recognition task was achieved by using a transfer learning approach from the image annotation task. Specifically, the VGG image embedding of a frame is projected using a linear transformation which was learned on matching images and sentences by the Canonical Correlation Analysis (CCA) algorithm~\cite{hotelling1936relations}.

The proposed RNN-FV method achieves state-of-the-art results in action recognition on the HMDB51~\cite{hmdb51} and UCF101~\cite{UCF101} datasets. In image annotation and image search tasks, the RNN-FV method is used for the representation of sentences and achieves state-of-the-art results on the Flickr8K dataset~\cite{hodosh2013framing} and competitive results on other benchmarks.

\section{Previous Work}
\label{sec:prev}

\paragraph{Action Recognition}
As in other object recognition problems, the standard pipeline in action recognition is comprised of three main steps: feature extraction, pooling and classification. Many works~\cite{HOHA08,wang2011action,kliper2012motion} have focused on the first step of extracting local descriptors. Laptev et al. ~\cite{STIP05} extend the notion of spatial interest points into the spatio-temporal domain and show how the resulting features can be used for a compact representation of video data. Wang et al.~\cite{rem_cordiiccv13,wang2013dense} used low-level hand-crafted features such as histogram of oriented gradients (HOG), histogram of optical flow (HOF) and motion boundary histogram (MBH).

Recent works have attempted to replace these hand-crafted features by deep-learned features for video action recognition due to its wide success in the image domain. Early attempts~\cite{taylor2010convolutional,ji20133d,karpathy2014large} achieved lower results in comparison to hand-crafted features, proving that it is challenging to apply deep-learning techniques on videos due to the relatively small number of available datasets and complex motion patterns. More recent attempts managed to overcome these challenges and achieve state of the art results with deep-learned features. Simonyan et al.~\cite{simonyan2014two} designed two-stream ConvNets for learning both the appearance of the video frame and the motion as reflected by the estimated optical flow. Du Tran et al.~\cite{tran2014learning} designed an effective approach for spatiotemporal feature learning using 3-dimensional ConvNets. 

In the second step of the pipeline, the pooling, Wang et al.~\cite{wang2013comparative} compared different pooling techniques for the application of action recognition and showed empirically that the Fisher Vector encoding has the best performance. Recently, more complex pooling methods were demonstrated by Peng et al.~\cite{peng2014action} who proposed Stacked Fisher Vectors (SFV), a multi-layer nested Fisher Vector encoding and Wang et al.~\cite{wang2015action} who proposed a trajectory-pooled deep-convolutional descriptor (TDD). TDD uses both a motion CNN, trained on UCF101, and an appearance CNN, originally trained on ImageNet~\cite{returnofdevil}, and fine-tuned on UCF101.

\paragraph{Image Annotation and Image Search} 

In the past few years, the state-of-the-art results in image annotation and image search have been provided by deep learning approaches~\cite{socher2013grounded,mikolajczyk2015deep,klein2015associating,karpathy2014deep,mao2014deep,kiros2014unifying,chen2014learning,Karpathy,vinyals2014show,ma2015multimodal}. A typical system is composed of three important components: (i) Image Representation, (ii) Sentence Representation, and (iii) Matching Images and Sentences. The image is usually represented by applying a pre-trained CNN on the image and taking the activations from the last hidden layer. 

There are several different approaches for the sentence representation; Socher et al.~\cite{socher2013grounded} used a dependency tree Recursive Neural Network. Yan et al.~\cite{mikolajczyk2015deep} used a TF-IDF histogram over the vocabulary. Klein et al.~\cite{klein2015associating} used word2vec~\cite{mikolov2013distributed} as the word embedding and then applied Fisher Vector based on a Hybrid Gaussian-Laplacian Mixture Model (HGLMM) in order to pool the word2vec embeddings of the words in a given sentence into a single representation.  Ma et al.~\cite{ma2015multimodal} proposed a matching CNN (m-CNN) that composes words to different semantic fragments and learns the inter-modal relations between image and the composed fragments at different levels.

Since a sentence can be seen as a sequence of words, many works have used a Recurrent Neural Network (RNN) in order to represent sentences~\cite{karpathy2014deep,vinyals2014show,mao2014deep,kiros2014unifying,kiros2015skip}. To address the need for capturing long term semantics in the sentence, these works mainly use Long Short-Term Memory (LSTM)~\cite{hochreiter1997long} or Gated Recurrent Unit (GRU)~\cite{chung2014empirical} cells. Generally, the RNN treats a sentence as an ordered sequence of words, and incrementally encodes a semantic vector of the sentence, word-by-word. At each time step, a new word is encoded into the semantic vector, until the end of the sentence is reached. All of the words and their dependencies will then have been embedded into the semantic vector, which can be used as a feature vector representation of the entire sentence. Our work also uses an RNN in order to represent sentences but takes the derived gradient from the RNN as features, instead of using a hidden or an output layer of the RNN.

A number of techniques have been proposed for the task of matching images and sentences. Klein et al.~\cite{klein2015associating} used CCA~\cite{hotelling1936relations} and Yan et al.~\cite{mikolajczyk2015deep} introduced a Deep CCA in order to project the images and sentences into a common space and then performed a nearest neighbor search between the images and the sentences in the common space. Kiros et al.~\cite{kiros2014unifying}, Karpathy et al.\cite{karpathy2014deep}, Socher et al.~\cite{socher2013grounded} and Ma et al.~\cite{ma2015multimodal} used a contrastive loss function trained on matching and unmatching pairs of (image,sentence) in order to learn a score function for a given pair. Mao et al.~\cite{mao2014deep} and Vinyals et al.~\cite{vinyals2014show} learned a probabilistic model for inferring a sentence given an image and, therefore, are able to compute the probability that a given sentence will be created by a given image and used it as the score.

\section{Baseline pooling methods}

In this section we describe two baseline pooling methods that can represent a multiset of vectors as a single vector. The notation of a multiset is used to clarify that the order of the words in a sentence does not affect the representation, and that a vector can appear more than once. Both methods can be applied to sequences, however, the resulting representation will be insensitive to ordering. To address this, we propose  in Sec.~\ref{sec:rnnfv} a novel pooling method: RNN-FV.

\subsection{Mean Vector}

This pooling technique takes a multiset of vectors, ${X=\{x_1,x_2,\dots,x_N\} \in R^D}$, and computes its mean: ${v=\frac{1}{N}\sum_{i=1}^{N} x_i}$.
Clearly, the vector $v$ that results from the pooling is in $R^D$.

The disadvantage of this method is the blurring of the multiset's content. Consider, for example, the text encoding task, where each word is represented by its word2vec embedding. By adding multiple vectors together, the location obtained -- in the semantic embedding space --  is somewhere in the convex hull of the words that belong to the multiset. 

\subsection{Fisher Vector of a GMM}

Given a multiset of vectors, $X=\{x_1,x_2,\dots,x_N\} \in R^D$, the standard FV~\cite{perronnin2007fisher} is defined as the gradient of the log-likelihood of $X$ with respect to the parameters of a pre-trained Diagonal-Covariance Gaussian Mixture Model (GMM). It is a common practice to limit the FV representation to the partial derivatives with respect to the means, $\mu$, and the standard deviations, $\sigma$, and ignore the partial derivatives with respect to the mixture weights.

It is worth noting the linear structure of the GMM FV pooling. Since the likelihood of the multiset is the multiplication of the likelihoods of the individual elements, the log-likelihood is additive. This convenient property would not be preserved in the RNN model, where the probability of an element in the sequence depends on all the previous elements.

To all types of FV, we apply the two improvements that were introduced by Perronnin et al.~\cite{perronnin2010large}. The first improvement is to apply an element-wise power normalization function, $f(z)=\textsf{sign}(z)|z|^{\alpha}$ where $0\leq\alpha\leq 1$ is a parameter of the normalization. The second improvement is to apply an L2 normalization on the FV after applying the power normalization function.

\section{RNN-Based Fisher Vector}
\label{sec:rnnfv}

The pooling methods described above share a common disadvantage: insensitivity to the order of the elements in the sequence. A way to tackle this, while keeping the power of gradient-based representation, would be to replace the Gaussian model by a generative sequence model that takes into account the order of elements in the sequence. A desirable property of the sequence model would be the ability to calculate the gradient (with respect to the model's parameters) of the likelihood estimate by this model to an input sequence.

In this section, we show that such a model can be obtained by training an RNN to predict the next element in a sequence, given the previous elements. Having this, we propose, for the first time, the RNN-FV: A Fisher Vector that is based on such an RNN sequence model.

We propose two types of RNN-FVs. One type is based on training a regression problem, and the other on training a classification problem. In practice, only the first type is directly useful for video analysis. For image annotation, the first type outperforms the second.

Given a sequence of vectors $S$ with $N$ vector elements $x_1,...,x_N$, we convert it to the input sequence $X=(x_0,x_1,...,x_{N-1})$, where $x_0 = x_{start}$. This special element is used to denote the beginning of the input sequence, and we use $x_{start} = 0$ throughout this paper. The RNN is trained to predict, at each time step $i$, the next element $x_{i+1}$ of the sequence, given the previous elements $x_0,...,x_i$. Therefore, given the input sequence, the target sequence would be: $Y=(x_1,x_2,...x_{N})$.

The training data and the training process are  application dependent, as is described in Sec.~\ref{sec:video} for action recognition and in Sec.~\ref{sec:lang} for image annotation.

\subsection{RNN Trained for Regression}
\label{sec:rnn_reg}

Given a sequence of input vectors $X$, the regression RNN is trained to predict the next vector in the sequence $S$, i.e., the sequence $Y$. The output layer of the network is a fully-connected layer, the size of which would be $D$, i.e., the dimension of the input vector space. 

There are several regression loss functions that can be used. Here, we consider the following loss function:
\begin{equation}
{Loss(y,v)=\frac12 \|y-v\|^2}
\end{equation}
where $y$ is the target vector and $v$ is the predicted vector.

After the RNN training is done, and given a new sequence $S$, the derived sequence $X$ is fed to the RNN. Denote the output of the RNN at time step $i$ $(i=0,...,N-1)$ by ${RNN(x_0,...,x_i)=v_i\in R^D}$. The target at time step $i$ is $x_{i+1}$ (the next element in the sequence), and the loss is:

\begin{equation}\label{eq:rnn_reg_loss_i}
{Loss(x_{i+1},v_i)=\frac12 \|x_{i+1}-v_i\|^2}
\end{equation}

The RNN can be seen as a generative model, and the likelihood of any vector $x$ being the next element of the sequence, given ${x_0,...,x_i}$, can be defined as:
\begin{equation}
p\left(x|x_0,...,x_i\right)=(2\pi)^{-D/2} \exp\left(-\frac12 \|x-v_i\|^2\right)
\end{equation}

We are generally interested in the likelihood of the correct prediction, i.e., in the likelihood of the  vector $x_{i+1}$ given ${x_0,...,x_i}$: $p\left(x_{i+1}|x_0,...,x_i\right)$.

The RNN-based likelihood of the entire sequence X is:
\begin{equation}
p(X)=\prod_{i=0}^{N-1} p\left(x_{i+1}|x_0,...,x_i\right)
\end{equation}

The negative log likelihood of $X$ is:
\begin{equation}\label{eq:rnn_reg_neg_log_likelihood}
\begin{split}
\mathcal{L}(X)&=-\log \left(p(X)\right)=-\sum_{i=0}^{N-1} \log \left(p\left(x_{i+1}|x_0,...,x_i\right) \right) \\
&=\frac{ND}{2} \log(2\pi)+\frac12\sum_{i=0}^{N-1} \|x_{i+1}-v_i\|^2
\end{split}
\end{equation}

In order to represent $X$ using the Fisher Vector scheme, we have to compute the gradient of $\mathcal{L}(X)$ with respect to our model's parameters. With RNN being our model, the parameters are the weights $W$ of the network. By \eqref{eq:rnn_reg_loss_i} and \eqref{eq:rnn_reg_neg_log_likelihood}, we get that $\mathcal{L}(X)$ equals the loss that would be obtained when $X$ is fed as input to the RNN, up to an additive constant. Therefore, the desired gradient can be computed by backpropagation: we feed $X$ to the network and perform forward and backward passes. The obtained gradient $\nabla_W \mathcal{L}(X)$ would be the (unnormalized) RNN-FV representation of $X$. Notice that this gradient is \textit{not} used to update the network's weights as done in training - here we perform backpropagation \textit{at inference time}.

Other loss functions may be used instead of the one presented in this analysis. Given a sequence, the gradient of the RNN loss may serve as the sequence representation, even if the loss is not interpretable as a likelihood.

\subsection{RNN Trained for Classification}
\label{sec:rnn_classifi}

The classification application is applicable for predicting a sequence of symbols $w_1$,$w_2$,...,$w_N$ that have matching vector representations $R(w_1) = x_1$, $R(w_2) = x_2$, ..., $R(w_N) = x_N$. The RNN predicts the sequence $U = (w_1 ,w_2 , \hdots ,w_N)$ from the sequence $X=(x_0,x_1,\hdots,x_{N-1})$.

Denote by $M$ the size of our symbol alphabet, i.e., the number of unique symbols in the input sequences. The output layer of the network is a softmax layer with $M$ units, where the $j$'th element in the output is the probability of the $j$'th symbol to be the next output element. The loss function for the training of the RNN is the cross-entropy loss.

After the RNN is trained, it is ready to be used as a feature vector extractor for new sequences. Denote the new sequence by $U$ and its vector representation by $X$ as above. Consider feeding the sequence $X$ to the RNN. At time step $i$ $(i=0,...,N-1)$, the output of the RNN is ${RNN(x_0,...,x_i)=(p_1^i,...,p_M^i)}$, where $\sum_{j=1}^{M} p_j^i=1$. Here, $p_j^i$ is the probability which the RNN gives to the $j$'th symbol at time step $i$.

The cross-entropy loss at time step $i$ is derived from the probability given to the correct next symbol:
\begin{equation}\label{eq:rnn_classif_loss_i}
loss_i=-\log \left(p_{w_{i+1}}^i\right)=-\log \left(\Pr\left(w_{i+1}|w_0,...,w_i\right)\right)
\end{equation}

The RNN can be seen as a generative model which gives likelihood to the sequence $U$:
\begin{equation}
\Pr(U)=\prod_{i=0}^{N-1} \Pr\left(w_{i+1}|w_0,...,w_i\right)=\prod_{i=0}^{N-1} p_{w_{i+1}}^i
\end{equation}

The negative log likelihood of $U$ is:
\begin{equation}\label{eq:rnn_classif_neg_log_likelihood}
\mathcal{L}(U)=-\log \left(\Pr(U)\right)=-\sum_{i=0}^{N-1} \log \left(p_{w_{i+1}}^i \right)
\end{equation}

By \eqref{eq:rnn_classif_loss_i} and \eqref{eq:rnn_classif_neg_log_likelihood}, we get that $\mathcal{L}(U)$ equals the loss that would be obtained when $X$ is fed as input, and $U$ as output to the RNN. Therefore, the desired gradient can be computed by backpropagation, i.e. feeding $X$ to the network and performing forward and backward passes. The obtained gradient $\nabla_W \mathcal{L}(U)$ would be the (unnormalized) RNN-FV representation of $U$.

\subsection{Normalization of the RNN-FV}
\label{sec:fim}

It was suggested by~\cite{perronnin2007fisher} that normalizing the FVs by the Fisher Information Matrix is beneficial. We approximated the diagonal of the Fisher Information Matrix (FIM), which is usually used for FV normalization. Note, however, that we did not observe any empirical improvement due to this normalization, and our experiments are reported without it.

Let $\omega \in W$ be a single weight of the RNN. The term in the diagonal of the FIM which corresponds to $\frac{\partial \mathcal{L}(X|W)}{\partial \omega }$ is: $F_{\omega }=\int_X p\left(X|W\right)\left[\frac{\partial \mathcal{L}\left(X|W \right)}{\partial \omega } \right] ^ {2} dX$.

Since the probabilistic model which determines $p\left(X|W\right)$ is the RNN, it is impossible to derive a closed-form expression for this term. Therefore, we approximated it directly from the gradients of the training sequences, by computing the mean of ${\left[\frac{\partial \mathcal{L}\left(X|W \right)}{\partial \omega } \right] ^ {2}}$ for each $\omega \in W$. The normalized partial derivatives of the FV are then: $F_{w}^{-1/2} \frac{\partial \mathcal{L}\left(X|W \right)}{\partial \omega }$.

\section{Action recognition pipeline}
\label{sec:video}

The action recognition pipeline contains the underlying appearance features used to encode the video, the sequence encoding using the RNN-FV, and an SVM classifier on top.

\subsection{Visual features}
\label{sec:videofeatures}

The RNN-FV is capable of encoding the sequence properties, and as underlying features, we rely on video encodings that are based on single frames or on fixed length blocks of frames.

\noindent{\bf VGG} Using the pre-trained VGG convolutional network~\cite{oxford}, we extract a 4096-dimensional representation of each video frame. The VGG pipeline is used, namely, the original image is cropped in ten different ways into 224 by 224 pixel images: the four corners, the center, and their x-axis mirror image. The mean intensity is then subtracted in each color channel and the resulting images are encoded by the network. The average of the 10 feature vectors obtained is then used as the single image representation. In order to speed up the method, the input video was sub-sampled, and one in every 10 frames was encoded. Empirically, we noticed that recognition performance was comparable to that of using all video frames. To further reduce run-time, the data dimensionality was reduced via PCA to 500D. In addition, L2 normalization was applied to each vector. All PCAs in this work were trained for each dataset and each training/test split separately, using only the training data.

\noindent{\bf CCA} Using the same VGG representation of video frames as mentioned above and the code of~\cite{klein2015associating}\footnote{Available at \url{www.cs.tau.ac.il/~wolf/code/hglmm}}, we represented each frame by a vector as follows: we considered the common image-sentence vector space obtained by the CCA algorithm, using the best model (GMM+HGLMM) of~\cite{klein2015associating} trained on the COCO dataset~\cite{coco}. We mapped each frame to that vector space, getting a 4096-dimensional image representation. As the final frame representation, we used the first (i.e.~the principal) 500 dimensions out of the 4096. For our application, the projected VGG representations were L2 normalized. The CCA was trained for an unrelated task of image to sentence matching, and its success, therefore, suggests a new application of transfer learning: from image annotation to action recognition. 

\noindent {\bf C3D} While the representations above encode single frames, the C3D method~\cite{tran2014learning} splits the video  into sub-volumes that are encoded one by one. Following the recommended settings, we applied the Du Tran et al. pre-trained 3D convolutional neural network in order to extract 4096D representation to 16-frame blocks. The blocks are sampled with an 8 frame stride. Following feature extraction, PCA dimensionality reduction (500D) and L2 normalization were applied.

\subsection{Network structure}

Our RNN model consists of three layers:  a 200D fully-connected layer units with Leaky-Relu activation  ($\alpha=0.1$), a 200-units Long Short-Term Memory (LSTM)~\cite{hochreiter1997long} layer, and a 500D linear fully-connected layer. Our network is trained for regression with the mean square error (MSE) loss function. Weight decay and dropouts were also applied. An improvement in recognition performance was noticed when the dropout rate was enlarged, up to a rate of 0.95, due to its ability to ensure the discriminative characteristics of each weight and hence also of each gradient. 

\subsection{Training and classification}
\label{sec:videornntraining}
We train the RNN to predict the next element in our video representation sequence, given the previous elements, as described in Sec.~\ref{sec:rnn_reg}. In our experiments, we use only the part of gradient corresponding to the weights of the last fully-connected layer. Empirically, we saw no improvement when using the partial derivatives with respect to weights of other layers. In order to obtain a fixed size representation, we average the gradients over all time steps. The gradient representation dimension is 500x201=100500, which is the number of weights in the last fully-connected layer. We then apply PCA to reduce the representation size to 1000D, followed by power and L2 normalization. 

Video classification is performed using a linear SVM with a parameter $C=1$. Empirically, we noticed that the the best recognition performance is obtained very quickly and hence early stopping is necessary. In order to choose an early stopping point we use a validation set. 
Some of the videos in the dataset are actually segments of the same original video, and are included in the dataset as different samples. Care was taken to ensure that no such similar videos are both in the training and validation sets, in order to guarantee that high validation accuracy will ensure good generalization and not merely over-fitting.

After each RNN epoch, we extract the RNN-FV representation as described above, train a linear SVM classifier on the training set and evaluate the performance on the validation set. The early stopping point is chosen at the  epoch with highest recognition accuracy on the validation set. After choosing our model this way, we train an SVM classifier on all training samples (training + validation samples) and report our performance on the test set. 

\section{Image-sentence retrieval}
\label{sec:lang}

In the image-sentence retrieval tasks, vector representations are extracted separately for the sentences and the images. These representations are then mapped into a common vector space, where the two are being matched. \cite{klein2015associating} have presented a similar pipeline for GMM-FV. We replace this representation with RNN-FV.

A sentence, being an ordered sequence of words, can be represented as a vector using the RNN-FV scheme. Given a sentence with $N$ words $w_1,...,w_N$, (where $w_N$ is considered to be the period, namely a $w_{end}$ special token), we treat the sentence as an ordered sequence $S=(w_0,w_1,...,w_{N-1})$, where $w_0 = w_{start}$. An RNN is trained to predict, at each time step $i$, the next word $w_{i+1}$ of the sentence, given the previous words $w_0,...,w_i$. Therefore, given the input sequence $S$, the target sequence would be: $(w_1,w_2,...w_N)$.

The training data may be any large set of sentences. These sentences may be extracted from the dataset of a specific benchmark, or, in order to obtain a generic representation, any external corpus, e.g., Wikipedia, may be used. 

The two network alternatives are explored: classification and regression. As observed in the action recognition case, we did not benefit from extracting partial derivatives with respect to the weights of the hidden layers, and hence we only use those of the output layer as our representation.

When the RNN is trained for classification, each word in the dictionary is considered as a class. The input to the network is the word's embedding, a 300D vector in our case. The hidden layer is LSTM with 512 units, which is followed by a softmax output layer. This design creates two challenges. The first is dimensionality: the size of the softmax layer is the size of the dictionary, $M$, which is typically large. As a result, $\nabla_W \mathcal{L}(X)$ has a high dimensionality. The second issue is with generalization capability: since the softmax layer is fixed, a network cannot handle a sentence containing a word that does not appear in its training data. 

When training the RNN for regression, the same 300D input is used, followed by an LSTM layer of size 100. The output layer, in this case, is fully-connected, where the (300 dimensional) word embedding of next word is predicted. We use no activation function at the output layer. Notice that the two issues pointed out regarding the classification RNN are not present in the regression case. First, the size $D$ of the output layer depends only on the dimension of the word embedding. Second, the network can naturally handle unseen words, since it predicts vectors in the word vector space rather than an index of a specific word.

For matching images and text, each image is represented as a 4096-dimensional vector extracted using the 19-layer VGG, as described in Sec.~\ref{sec:videofeatures}. The regularized CCA algorithm~\cite{Vinod1976147}, where the regularization parameter is selected based on the validation set, is used to match the the VGG representation with the sentence RNN-FV representation. In the shared CCA space, the cosine similarity is used.

We explored several configurations for training the RNN. 
\noindent {\bf RNN training data} We employed either the training data of each split in the respective benchmark, or the 2010-English-Wikipedia-1M dataset made available by the Leipzig Corpora Collection~\cite{quasthoff2006corpus}. This dataset contains 1 million sentences randomly sampled from English Wikipedia.
\noindent {\bf  Word embedding} A word was represented either by word2vec, or by the GMM+HGLMM representation of~\cite{klein2015associating}, projected to a 300D sentence to VGG-encoded-image CCA space. We made sure to match the training split according to the benchmark tested. 
\noindent {\bf  Sentence sequence direction} We explored both the conventional left-to-right sequence of words and the reverse direction.

\section{Experiments}

We evaluated the effectiveness of the various pooling methods on two important yet distinct application domains: action recognition and image textual annotation and search. 

As mentioned, applying the FIM normalization (Sec.~\ref{sec:fim}) did not seem to improve results. Another form of normalization we have tried, is to normalize each dimension of the gradient by subtracting its mean and dividing by its standard deviation. This also did not lead to an improved performance. Two normalizations that were found to be useful are the Power Normalization and the L2 Normalization, which were introduced in~\cite{perronnin2010improving} (see Section~\ref{sec:prev}). Both are employed, using a constant $\alpha=1/2$.

\subsection{Action recognition}

\begin{table*}[t]
\begin{minipage}[c]{0.70\textwidth}
\begin{center}
\begin{tabular}{|l|c|c|c|c|c|c|}
\hline
Dataset        &      \multicolumn{3}{|c|}{{HMDB51}} &        \multicolumn{3}{|c|}{{UCF101}}        \\ \hline
Method & MP & GMM-FV          & RNN-FV & MP & GMMFV          & RNN-FV \\ \hline
 \hline
VGG PCA        & 42.16        & 36.8           & 45.62   & 75.51        & 76.53           & 79.29  \\ \hline
VGG CCA        & 43.05        & 39.61           & 46.14  & 77.49        & 76.84           & 79.49  \\ \hline
C3D            & 51.2        & 45.82          & 52.88  & 81.05        & 80.04           & 82.33  \\ \hline
\end{tabular}
\end{center}
\end{minipage}\hfill
\begin{minipage}[c]{0.3\textwidth}
\caption{Comparing pooling techniques (mean pooling, GMM-FV and RNN-FV) on HMDB51 and UCF101. Three types of features are used: VGG-PCA, VGG-CCA, and C3D. The table reports recognition average accuracy (higher is better).}
\label{tab:comp_pool}
\end{minipage}
\end{table*}

Our experiments were conducted on two large action recognition benchmarks. The UCF101~\cite{UCF101} dataset consists of 13,320 realistic action videos, collected from YouTube, and divided into 101 action categories. We use the three splits provided with this dataset in order to evaluate our results and report the mean average accuracy over these splits.

The HMDB51 dataset~\cite{hmdb51} consists of 6766 action videos, collected from various sources, and divided into 51 action categories. Three splits are provided as an official benchmark and are used here. The mean average accuracy over these splits is reported.

Table~\ref{tab:comp_pool} compares our RNN-FV pooling method to Mean and GMM-FV pooling. Three sets of features, as described in Sec.~\ref{sec:videofeatures} are used: VGG coupled with PCA, VGG projected by the image to sentence matching CCA, and C3D. 

The parameters were set on the validation split that we created for the provided training set. For GMM-FV, the only parameter is $k$, which is the number of components in the mixture. The validated values of $k$ were in the set $\{1,2,4,8,16,32\}$. The parameter for RNN-FV was the stopping point of the RNN training, as described in Sec.~\ref{sec:videornntraining}. Classification is conducted in all experiments using a multiclass (one-vs-all) linear SVM with C=1. 

As can be seen in table~\ref{tab:comp_pool}, the RNN-FV pooling outperformed the other pooling methods by a sizable margin. Another interesting observation is that with VGG frame representation, CCA outperformed PCA consistently in all pooling methods. Not shown is the performance obtained when using the activations of the RNN as a feature vector. These results are considerably worse than all pooling methods. Notice that the representation dimension of Mean pooling is 500 (like the features we used), the GMM-FV dimension is $2\times k\times 500$, where k is the number of clusters and the RNN-FV dimension is 1000.

Table~\ref{tab:comp_sota} compares our proposed RNN-FV method, combining multiple features together, with recently published methods on both datasets. The combinations were performed using early fusion, i.e, we concatenated the normalized low-dimensional gradients of the models and train multi-class linear SVM on the combined representation. We also tested the combination of our two best models with idt ~\cite{wang2013action} and got state of the art results on both benchmarks.  Interestingly, when training the RNNs on UCF101 and applying to encode HMDB51 videos, a comparable results of $66.99$ ($54.47$ without idt) is obtained, which is also above current state of the art.
\begin{table}[]
\begin{center}
\begin{tabular}{|l|c|c|}
\hline
Method                           & HMDB51 & UCF101 \\ \hline \hline
idt~\cite{wang2013action}                              & 57.2   & 85.9   \\ \hline
idt + high-D encodings~\cite{peng2014bag} & 61.1   & 87.9   \\ \hline
Two-stream CNN (2 nets)~\cite{simonyan2014two}                   & 59.4   & 88     \\ \hline
Multi-skip Feature Stacking~\cite{lan2014beyond}      & 65.4   & 89.1   \\ \hline
C3D (1 net)~\cite{tran2014learning}                      & --      & 82.3   \\ 
C3D (3 nets)~\cite{tran2014learning}                     & --     & 85.2   \\ 
C3D (3 nets) + idt~\cite{tran2014learning}               & --     & 90.4   \\ \hline
TDD (2 nets)~\cite{wang2015action}                              & 63.2   & 90.3   \\ 
TDD (2 nets) +  idt~\cite{wang2015action}                       & 65.9   & 91.5   \\ \hline
stacked FV~\cite{peng2014action}                       & 56.21  & --     \\ 
stacked FV + idt~\cite{peng2014action}                       & 66.78  & --     \\ \hline
RNN-FV(C3D + VGG-CCA)                & 54.33  & 88.01  \\
RNN-FV(C3D + VGG-CCA) + idt          & {\bf 67.71}  & {\bf 94.08}  \\ \hline
\end{tabular}
\end{center}
\caption{comparison to the state of the art on UCF101 and HMDB51. In order to obtain the best performance, we combine, similar to all other contributions, multiple features. We also present a result where idt~\cite{wang2013action} is combined, similar to all other top results (Multi-skip extends idt). This adds motion based information to our method.}
\label{tab:comp_sota}
\end{table}

\subsection{Image-sentence retrieval}

The effectiveness of RNN-FV as sentence representation is evaluated on the bidirectional image and sentence retrieval task. We perform our experiments on three benchmarks: Flickr8K~\cite{hodosh2013framing},  Flickr30K~\cite{hodoshimage}, and COCO~\cite{coco}. The datasets contain $8,000$, $30,000$, and $123,000$ images respectively. Each image is accompanied with 5 sentences describing the image content, collected via crowdsourcing.

The Flickr8k dataset is provided with training, validation, and test splits. For Flickr30K and COCO, no training splits are given, and we use the same splits used by~\cite{klein2015associating}.

There are three tasks in this benchmark: image annotation, in which the goal is to retrieve, given a query image, the five ground truth sentences; image search, in which, given a query sentence, the goal is to retrieve the ground truth image; and sentence similarity, in which the goal is, given a sentence, to retrieve the other four sentences describing the same image. Evaluation is performed using Recall@K, namely the fraction of times the correct result was ranked within the top K items. The median and mean rank of the first ground truth result are also reported. For the sentence similarity task, only mean rank is reported.


As mentioned in Sec.~\ref{sec:lang}, we explored RNN-FV based on several RNNs. The first RNN is a generic one: it was trained with the Wikipedia sentences as training data and word2vec as word embedding. In addition, for each of the three datasets, we trained three RNNs with the dataset's training sentences as training data: one with word2vec as word embedding; one with the "CCA word embedding" derived from the semantic vector space of~\cite{klein2015associating}, as explained in Sec.~\ref{sec:lang}; and one with the CCA word embedding, and with feeding the sentences in reverse order. These RNNs were all trained for regression. For Flickr8K, we also trained an RNN for classification (with Flickr8K training sentences, and word2vec embedding). In this network, the softmax layer was of size 8,148, corresponding to the number of unique words in the Flickr8k dataset. Since the resulting number of weights of the output layer is around 4 million, we reduced the dimension of the gradient feature vector by random sampling of 72,000 coordinates. Training a classification model on the larger datasets is virtually impractical, since the number of unique words in these datasets is much higher, resulting in a very large softmax layer and a huge number of weights. 

In the regression RNNs, we used an LSTM layer of size 100. We did not observe a benefit in using more LSTM units. We used the part of the gradient corresponding to all 30,300 weights of the output layer (including one bias per word2Vec dimension). In the case of the larger COCO dataset, due to the computational burden of the CCA calculation, we used PCA to reduce the gradient dimension from 30,300 to 20,000. PCA was calculated on a random subset of 300,000 sentences (around 50\%) of the training set. We also tried PCA dimension reduction to a lower dimension of 4,096, for all three datasets. We observed no change in performance (Flickr8K) or slightly worse results (Flickr30K and COCO).

The number of RNN training epochs was 400, 100, 20, and 15, for the Flickr8k, Flickr30k, COCO and Wikipedia datasets respectively.

\begin{table*}[t]
\centering
\begin{tabular}{|l|lllll|lllll|l|}
\hline
& \multicolumn{5}{|c|}{Image Annotation} & \multicolumn{5}{|c|}{Image Search} & {Sentence}\\
& r@1 & r@5 & r@10 & median & mean & r@1 & r@5 & r@10 & median & mean & mean \\
& & & & rank & rank & & & & rank & rank& rank\\
\hline

SDT-RNN~\cite{socher2013grounded}            & 6.0  & 22.7 & 34.0 & 23.0 & NA   & 6.6  & 21.6 & 31.7 & 25.0 & NA   & NA   \\
DFE~\cite{karpathy2014deep}                  & 12.6 & 32.9 & 44.0 & 14.0 & NA   & 9.7  & 29.6 & 42.5 & 15.0 & NA   & NA   \\
RVP~\cite{chen2014learning}                  & 11.7 & 34.8 & 48.6 & 11.2 & NA   & 11.4 & 32.0 & 46.2 & 11.0 & NA   & NA   \\
DVSA~\cite{Karpathy}                         & 16.5 & 40.6 & 54.2 & 7.6  & NA   & 11.8 & 32.1 & 44.7 & 12.4 & NA   & NA   \\
SC-NLM~\cite{kiros2014unifying}              & 18.0 & 40.9 & 55.0 & 8.0  & NA   & 12.5 & 37.0 & 51.5 & 10.0 & NA   & NA   \\
DCCA~\cite{mikolajczyk2015deep}              & 17.9 & 40.3 & 51.9 & 9.0  & NA   & 12.7 & 31.2 & 44.1 & 13.0 & NA   & NA   \\
NIC~\cite{vinyals2014show}                   & 20.0 & NA   & 61.0 & 6.0  & NA   & 19.0 & NA   & 64.0 & {\bf 5.0}  & NA   & NA   \\
m-RNN~\cite{mao2014explain}                  & 14.5 & 37.2 & 48.5 & 11.0 & NA   & 11.5 & 31.0 & 42.4 & 15.0 & NA   & NA   \\
m-CNN~\cite{ma2015multimodal}                & 24.8 & 53.7 & 67.1 & 5.0  & NA   & 20.3 & 47.6 & 61.7 & {\bf 5.0}  & NA   & NA   \\
MeanVector~\cite{klein2015associating}       & 22.6 & 48.8 & 61.2 & 6.0  & 28.7 & 19.1 & 45.3 & 60.4 & 7.0  & 27.0 & 12.5 \\
GMM-FV~\cite{klein2015associating}           & 28.4 & 57.7 & 70.1 & 4.0  & 20.1 & 20.6 & 48.6 & 64.2 & 6.0  & 21.8 & 10.8 \\
MM-ENS~\cite{klein2015associating}           & 31.0 & 59.3 & 73.7 & 4.0  & 18.4 & 21.3 & 50.1 & 64.8 & {\bf 5.0}  & 21.0 & 10.5 \\
\hline
Our RNN-FV:                                    &      &      &      &      &      &      &      &      &      &      &      \\
wiki,w2v                                       & 29.3 & 57.8 & 70.8 & 4.0  & 21.4 & 19.8 & 48.5 & 62.9 & 6.0  & 25.2 & 10.0 \\
w2v                                            & 27.4 & 57.9 & 70.5 & 4.0  & 22.7 & 20.4 & 49.1 & 63.4 & 6.0  & 25.5 & 10.4 \\
w2v,clsf                                       & 28.3 & 57.2 & 69.8 & 4.0  & 19.9 & 20.0 & 47.8 & 62.8 & 6.0  & 27.0 & 13.2 \\
cca                                            & 30.9 & 60.1 & 73.1 & 4.0  & 19.4 & 20.7 & 48.7 & 63.8 & 6.0  & 29.2 & 11.3 \\
cca,rvrs                                       & 29.1 & 57.3 & 71.7 & 4.0  & 18.4 & 20.8 & 48.5 & 62.9 & 6.0  & 30.2 & 12.5 \\
cca + rvrs                                       & 30.8 & 59.8 & 72.9 & 4.0  & 18.2 & 21.8 & 49.6 & 64.4 & 6.0  & 27.3 & 11.2 \\
cca +~\cite{klein2015associating}             & {\bf 32.9} & {\bf 61.7} & {\bf 74.9} & {\bf 3.0}  & 16.8 & 22.0 & 51.5 & 66.5 & {\bf 5.0}  & 20.7 & 9.4  \\
cca + rvrs +~\cite{klein2015associating}        & 32.1 & 60.7 & 74.8 & {\bf 3.0}  & {\bf 16.5} & 22.1 & 51.4 & 66.5 & {\bf 5.0}  & 21.4 & 9.5  \\
all rnn-fv models                                & 29.9 & 60.7 & 73.4 & 4.0  & 17.9 & 22.4 & 52.7 & 67.2 & {\bf 5.0}  & 20.9 & 8.7  \\
all rnn-fv models +~\cite{klein2015associating} & 31.6 & 61.2 & 74.3 & {\bf 3.0}  & 17.4 & {\bf 23.2} & {\bf 53.3} & {\bf 67.8} & {\bf 5.0}  & {\bf 19.4} & {\bf 8.5} \\

\hline
\end{tabular}
\vspace{.2cm}
\caption{Image annotation, image search and sentence similarity results on the Flickr8k dataset. Shown are the recall rates at $1$, $5$, and $10$ retrieval results (higher is better). Also shown are the median and mean rank of the first ground truth (lower is better). We compare the results of previous work to variants of our RNN-FV. The `wiki' notation indicates that the RNN was trained on Wikipedia and not on the sentences of the specific dataset. `clsf' uses classification-RNN, while the other models were trained for regression. Models notated by `w2v' employ word2vec, while the other models (`cca') use the CCA embedding of~\cite{klein2015associating}. `rvrs' models were trained on reversed sentences. We also report results of combinations: `cca' and `reverse' models; `cca' and the best model (GMM+HGLMM) of~\cite{klein2015associating} (`MM-ENS'); `cca', `reverse' and~\cite{klein2015associating}; All RNN-FV models; All RNN-FV models and~\cite{klein2015associating}.
}
\label{tab:f8k}
\end{table*}

\begin{table*}[t]
\centering
\begin{tabular}{|l|lllll|lllll|l|}
\hline
& \multicolumn{5}{|c|}{Image Annotation} & \multicolumn{5}{|c|}{Image Search} & {Sentence}\\
& r@1 & r@5 & r@10 & median & mean & r@1 & r@5 & r@10 & median & mean & mean \\
& & & & rank & rank & & & & rank & rank& rank\\
\hline

SDT-RNN~\cite{socher2013grounded}            & 9.6  & 29.8 & 41.1 & 16.0 & NA   & 8.9  & 29.8 & 41.1 & 16.0 & NA   & NA   \\
DFE~\cite{karpathy2014deep}                  & 14.2 & 37.7 & 51.3 & 10.0 & NA   & 10.2 & 30.8 & 44.2 & 14.0 & NA   & NA   \\
RVP~\cite{chen2014learning}                  & 12.1 & 27.8 & 47.8 & 11.0 & NA   & 12.7 & 33.1 & 44.9 & 12.5 & NA   & NA   \\
DVSA~\cite{Karpathy}                         & 22.2 & 48.2 & 61.4 & 4.8  & NA   & 15.2 & 37.7 & 50.5 & 9.2  & NA   & NA   \\
SC-NLM~\cite{kiros2014unifying}              & 23.0 & 50.7 & 62.9 & 5.0  & NA   & 16.8 & 42.0 & 56.5 & 8.0  & NA   & NA   \\
DCCA~\cite{mikolajczyk2015deep}              & 16.7 & 39.3 & 52.9 & 8.0  & NA   & 12.6 & 31.0 & 43.0 & 15.0 & NA   & NA   \\
NIC~\cite{vinyals2014show}                   & 17.0 & NA   & 56.0 & 7.0  & NA   & 17.0 & NA   & 57.0 & 7.0  & NA   & NA   \\
LRCN~\cite{donahue2014long}                  & NA   & NA   & NA   & NA   & NA   & 17.5 & 40.3 & 50.8 & 9.0  & NA   & NA   \\
RTP~\cite{plummer2015flickr30k}(manual annotations)  & {\bf 37.4} & 63.1 & 74.3 & NA   & NA   & 26.0 & 56.0 & 69.3 & NA   & NA   & NA   \\
m-RNN~\cite{mao2014explain}                  & 35.4 & 63.8 & 73.7 & {\bf 3.0}  & NA   & 22.8 & 50.7 & 63.1 & 5.0  & NA   & NA   \\
m-CNN~\cite{ma2015multimodal}                & 33.6 & {\bf 64.1} & {\bf 74.9} & {\bf 3.0}  & NA   & 26.2 & {\bf 56.3} & 69.6 & {\bf 4.0}  & NA   & NA   \\
MeanVector~\cite{klein2015associating}       & 24.9 & 52.5 & 64.4 & 5.0  & 27.3 & 20.5 & 46.4 & 59.3 & 6.8  & 32.3 & 16.2 \\
GMM-FV~\cite{klein2015associating}           & 33.0 & 60.8 & 72.0 & {\bf 3.0}  & 19.0 & 23.9 & 51.7 & 64.9 & 5.0  & 24.8 & 15.0 \\
MM-ENS~\cite{klein2015associating}           & 35.0 & 62.1 & 73.8 & {\bf 3.0}  & 17.4 & 25.1 & 52.8 & 66.1 & 5.0  & 23.7 & 14.1 \\
\hline
Our RNN-FV:                                    &      &      &      &      &      &      &      &      &      &      &      \\
wiki,w2v                                       & 32.9 & 59.6 & 72.1 & {\bf 3.0}  & 18.5 & 23.9 & 52.0 & 65.2 & 5.0  & 26.0 & 15.2 \\
w2v                                            & 32.0 & 59.5 & 71.4 & {\bf 3.0}  & 17.2 & 23.4 & 51.7 & 65.2 & 5.0  & 24.5 & 14.1 \\
cca                                            & 33.6 & 60.5 & 73.0 & {\bf 3.0}  & 15.7 & 24.5 & 52.5 & 66.3 & 5.0  & 27.7 & 16.9 \\
cca,rvrs                                       & 32.8 & 61.9 & 72.7 & {\bf 3.0}  & 17.4 & 24.4 & 51.2 & 64.6 & 5.0  & 28.9 & 16.1 \\
cca + rvrs                                       & 33.6 & 62.4 & 73.4 & {\bf 3.0}  & 15.5 & 25.0 & 53.6 & 66.9 & 5.0  & 26.2 & 15.5 \\
cca +~\cite{klein2015associating}             & 35.1 & 63.3 & 74.2 & {\bf 3.0}  & 15.3 & 26.4 & 54.9 & 68.6 & {\bf 4.0}  & 21.7 & 13.4 \\
cca + rvrs +~\cite{klein2015associating}        & 35.1 & 63.5 & 74.5 & {\bf 3.0}  & {\bf 15.0} & 26.5 & 55.2 & 68.5 & {\bf 4.0}  & 22.0 & 13.5 \\
all rnn-fv models                                & 34.7 & 62.7 & 72.6 & {\bf 3.0}  & 15.6 & 26.2 & 55.1 & 69.2 & {\bf 4.0}  & 21.2 & 12.8 \\
all rnn-fv models +~\cite{klein2015associating} & 35.6 & 62.5 & 74.2 & {\bf 3.0}  & {\bf 15.0} & {\bf 27.4} & 55.9 & {\bf 70.0} & {\bf 4.0}  & {\bf 20.0} & {\bf 12.2} \\

\hline
\end{tabular}
\vspace{.2cm}
\caption{Image annotation, image search and sentence similarity results on the Flickr30k dataset. For details, see Table~\ref{tab:f8k}. The RTP method~\cite{plummer2015flickr30k} enjoys additional information that is not accessible to the other methods:  manual annotations of bounding boxes in the images, which were collected via crowdsourcing.
}
\label{tab:f30k}
\end{table*}

\begin{table*}[t]
\centering
\begin{tabular}{|l|lllll|lllll|l|}
\hline
& \multicolumn{5}{|c|}{Image Annotation} & \multicolumn{5}{|c|}{Image Search} & {Sentence}\\
& r@1 & r@5 & r@10 & median & mean & r@1 & r@5 & r@10 & median & mean & mean \\
& & & & rank & rank & & & & rank & rank& rank\\
\hline

DVSA~\cite{Karpathy}                         & 38.4 & 69.9 & 80.5 & {\bf 1.0} & NA   & 27.4 & 60.2 & 74.8 & {\bf 3.0} & NA   & NA   \\
m-RNN~\cite{mao2014explain}                  & 41.0 & 73.0 & 83.5 & 2.0 & NA   & 29.0 & 42.2 & 77.0 & {\bf 3.0} & NA   & NA   \\
m-CNN~\cite{ma2015multimodal}                & {\bf 42.8} & 73.1 & 84.1 & 2.0 & NA   & {\bf 32.6} & {\bf 68.6} & {\bf 82.8} & {\bf 3.0} & NA   & NA   \\
STV~\cite{kiros2015skip}                     & 33.8 & 67.7 & 82.1 & 3.0 & NA   & 25.9 & 60.0 & 74.6 & 4.0 & NA   & NA   \\
MeanVector~\cite{klein2015associating}       & 33.2 & 61.8 & 75.1 & 3.0 & 14.5 & 24.2 & 56.4 & 72.4 & 4.0 & 14.7 & 14.3 \\
GMM-FV~\cite{klein2015associating}           & 39.0 & 67.0 & 80.3 & 3.0 & 11.2 & 24.2 & 59.3 & 76.0 & 4.0 & 11.3 & 12.4 \\
MM-ENS~\cite{klein2015associating}           & 39.4 & 67.9 & 80.9 & 2.0 & 10.4 & 25.2 & 59.9 & 76.7 & 4.0 & 11.0 & 12.9 \\
\hline
Our RNN-FV:                                    &      &      &      &     &      &      &      &      &     &      &      \\
wiki,w2v                                       & 37.7 & 70.5 & 81.0 & 2.0 & 9.9  & 26.6 & 61.1 & 76.9 & 4.0 & 10.9 & 11.9 \\
w2v                                            & 39.9 & 71.5 & 81.3 & 2.0 & 10.5 & 26.9 & 61.8 & 77.4 & 4.0 & 11.4 & 12.1 \\
cca                                            & 40.9 & {\bf 75.0} & {\bf 84.9} & 2.0 & 8.2  & 30.2 & 65.0 & 80.4 & {\bf 3.0} & 11.1 & 13.2 \\
cca,rvrs                                       & 41.3 & 71.5 & 83.7 & 2.0 & {\bf 8.1}  & 28.9 & 64.5 & 79.9 & {\bf 3.0} & 11.3 & 12.6 \\
cca + rvrs                                       & 40.8 & 73.4 & 84.1 & 2.0 & 8.2  & 30.4 & 65.5 & 80.9 & {\bf 3.0} & 10.7 & 12.3 \\
cca +~\cite{klein2015associating}             & 40.7 & 72.3 & 83.5 & 2.0 & 9.1  & 28.1 & 64.1 & 79.8 & {\bf 3.0} & 10.2 & 11.5 \\
cca + rvrs +~\cite{klein2015associating}        & 40.2 & 72.7 & 84.2 & 2.0 & 8.6  & 29.0 & 64.8 & 80.2 & {\bf 3.0} & 10.1 & 11.5 \\
all rnn-fv models                                & 40.8 & 71.9 & 83.2 & 2.0 & 8.9  & 29.6 & 64.8 & 80.5 & {\bf 3.0} & 9.7  & 10.6 \\
all rnn-fv models +~\cite{klein2015associating} & 41.5 & 72.0 & 82.9 & 2.0 & 9.0  & 29.2 & 64.7 & 80.4 & {\bf 3.0} & {\bf 9.5}  & {\bf 10.2} \\

\hline
\end{tabular}
\vspace{.2cm}
\caption{Image annotation, image search and sentence similarity results on the COCO dataset. For details, see Table~\ref{tab:f8k}
}
\label{tab:coco}
\end{table*}

Tables~\ref{tab:f8k},~\ref{tab:f30k} and~\ref{tab:coco} show the results of the different RNN-FV variants compared to the current state of the art methods. We also report results of combinations of models. Combining was done by averaging the image-sentence (or sentence-sentence) cosine similarities obtained by each model.

First, we see that regression-based RNN-FV should be preferred over the classification-based one. In addition to its lower dimension and natural handling of unseen words, the results obtained by regression RNN-FV are better. Second, we notice the competitive performance of the model trained on Wikipedia sentences, which demonstrates the generalization power of the RNN-FV, being able to perform well on data different than the one which the RNN was trained on. Training using the dataset's sentences only slightly improves result, and not always. Improved results are obtained when using the CCA word embedding instead of word2vec. It is interesting to see the result of the ``reverse'' model, which is on a par with the other models. It is somewhat complementary to the ``left-to-right'' model, as the combination of the two yields somewhat improved results. Finally, the combination of RNN-FV with the best model (GMM+HGLMM) of~\cite{klein2015associating} outperforms the current state of the art on Flickr8k, and is competitive on the other datasets.

\section{Conclusions}

This paper introduces a novel FV representation for sequences that is derived from RNNs. The proposed representation is sensitive to the element ordering in the sequence and provides a richer model than the additive ``bag'' model typically used for conventional FVs.

The RNN-FV representation surpasses the state-of-the-art results for video action recognition on two challenging datasets. When used for representing sentences, the RNN-FV representation achieves state-of-the-art or competitive results on image annotation and image search tasks. Since the length of the sentences in these tasks is usually short and, therefore, the ordering is less crucial, we believe that using the RNN-FV representation for tasks that use longer text will provide an even larger gap between the conventional FV and the RNN-FV.

A transfer learning result from the image annotation task to the video action recognition task was shown. The conceptual distance between these two tasks makes this result both interesting and surprising. It supports a human development-like way of training, in which visual labeling is learned through natural language, as opposed to, e.g., associating bounding boxes with nouns. While such training was used in computer vision to learn related image to text tasks, and while recently zero-shot action recognition was shown~\cite{JainICCV15,DBLP:journals/corr/XuHG15}, NLP to video action recognition transfer was never shown to be as general as presented here.

\section*{Acknowledgments}
This research is supported by the Intel Collaborative Research Institute for Computational Intelligence (ICRI-CI).

{\small
\bibliographystyle{ieee}
\bibliography{all_bib}
}

\end{document}